\documentclass[preprint]{IEEEtran}
\IEEEoverridecommandlockouts
\usepackage{cite}
\usepackage{amsmath,amssymb,amsfonts}
\usepackage{algorithmic}
\usepackage{graphicx}
\usepackage{textcomp}
\usepackage{xcolor}
\usepackage{hyperref}
\usepackage{subfig}
\usepackage{booktabs} 
\usepackage{balance}
\usepackage{colortbl}
\usepackage{pgfplots, pgfplotstable}
\pgfplotsset{compat=1.14}
\begin{document}

\title{Identifying Surgical Instruments in Laparoscopy Using Deep Learning Instance Segmentation}

\author{\IEEEauthorblockN{Sabrina Kletz, Klaus Schoeffmann}
\IEEEauthorblockA{\textit{Klagenfurt University, Austria} \\
sabrina@itec.aau.at, ks@itec.aau.at}
\and
\IEEEauthorblockN{Jenny Benois-Pineau}
\IEEEauthorblockA{\textit{University of Bordeaux, France} \\
jenny.benois@labri.fr} 
\and
\IEEEauthorblockN{Heinrich Husslein}
\IEEEauthorblockA{\textit{Medical University of Vienna, Austria} \\
heinrich.husslein@meduniwien.ac.at}
}

\IEEEoverridecommandlockouts
\IEEEpubid{\makebox[\columnwidth]{978-1-7281-4673-7/19/\$31.00~\copyright2019 IEEE \hfill} \hspace{\columnsep}\makebox[\columnwidth]{ }}

\maketitle

\IEEEpubidadjcol

\begin{abstract}
Recorded videos from surgeries have become an increasingly important information source for the field of medical endoscopy, since the recorded footage shows every single detail of the surgery.  However, while video recording is straightforward these days, automatic content indexing -- the basis for content-based search in a medical video archive -- is still a great challenge due to the very special video content. In this work, we investigate segmentation and recognition of surgical instruments in videos recorded from laparoscopic gynecology. More precisely, we evaluate the achievable performance of segmenting surgical instruments from their background by using a region-based fully convolutional network for instance-aware (1) instrument segmentation as well as (2) instrument recognition. While the first part addresses only binary segmentation of instances (i.e., distinguishing between instrument or background) we also investigate multi-class instrument recognition (i.e., identifying the type of instrument). Our evaluation results show that even with a moderately low number of training examples, we are able to localize and segment instrument regions with a pretty high accuracy. However, the results also reveal that determining the particular instrument is still very challenging, due to the inherently high similarity of surgical instruments.  
\end{abstract}

\begin{IEEEkeywords}
surgical instruments, laparoscopic videos, instance segmentation, deep learning, data augmentation, region-based convolutional neural network
\end{IEEEkeywords}

\section{Introduction}
\label{sec:introduction}
Laparoscopy is a minimally-invasive surgery that is performed via three to four orifices in the human body, where one is used for the laparoscope and the remaining ones for operation instruments. The laparoscope is a thin tool with a tiny high-resolution camera at the end, whose images are transmitted to a monitor in the operation room, allowing the surgeons to perform controlled and supervised actions. The video signal from the laparoscope also enables operating surgeons to record their procedures for retrospective analysis and post-operative \textit{Surgical Quality Assessment} (SQA)~\cite{Husslein2015}. For SQA, video recordings are closely investigated after the operation in order to assess the surgeon's skill level and find potential technical errors. In particular, the handling of instruments is observed, thus, surgeons are highly interested in the automated identification of instruments in videos. However, this would not only be beneficial for surgeons performing the assessment process, but also for further automated analysis techniques, since instruments are the region of interest, which in turn could serve as input for other analysis methods. 

In recent years, some efforts have been made to address the automated detection, segmentation, and tracking of instruments in minimally-invasive surgery. However, majority of works deal with the task of identifying instruments in robotic surgery \cite{Lee2019SegmentationRI,Shvets2018SegmentationRIEndovis17,Garcia2017ToolNet,Pakhomov2017RIResNetFCN,Laina2017,Garcia2017SIFCN}. In fields like laparoscopic cholecystectomy, on the other hand, automated tool presence detection has got attention \cite{Twinanda2016, Jin2018} because this is beneficial for surgical workflow analysis. One reason for the lack in research on instrument segmentation in traditional laparoscopy is that no datasets are provided, whereas in the field of robotic surgery one dataset is available, which has been released together with an endoscopic vision challenge~\cite{Allan2017EndoVis}. Although instruments in traditional laparoscopy strongly differ from those in robotic surgery, researchers in both application domains are facing similar challenges because videos, recorded in the inner of the abdomen, are accompanied with reflections, blurriness and smoke, which in turn causes difficulties for automatic content analysis. 
However, works that process robotic surgery videos have shown that using deep learning approaches are more robust against such noise in comparison to traditional approaches~\cite{Letouzey2016,Allan2017EndoVis}, where specific visual features have to be selected manually to describe the task at hand. 

We consider the task of identifying instruments as multi-class instance segmentation problem and investigate to what extent instruments can be localized, classified and separated from its background. In doing so, we examine a deep learning instance segmentation approach and compare it for two instrument segmentation scenarios: instance identification only, and multi-class instance recognition. To differentiate between them, we introduce the terms `\textit{binary instrument segmentation}' to segment each instrument separately and `\textit{multi-class instrument recognition}' to further distinguish between their types. We are interested in determining the precision of detection and classification results for both tasks the binary instrument segmentation and the multi-class instrument segmentation and recognition. For studying these tasks, we introduce a dataset for instrument segmentation in laparoscopy comprising instruments specifically employed in gynecologic myomectomy and hysterectomy. This dataset consists of extracted frames from various surgery videos and manually annotated segmentation masks for each visible instrument instance in one frame.

\section{Related Works}
Instrument recognition in medical endoscopy has been addressed in literature by using different approaches based on deep learning. These works can be categorized according to their approaches: semantic segmentation and tracking of instruments in robotic surgery \cite{Laina2017, Garcia2017ToolNet,Shvets2018SegmentationRIEndovis17,Pakhomov2017RIResNetFCN, Hasan2019UNetPlus} and instrument detection in cholecystectomy \cite{Twinanda2016, Jin2018}. However, this issue of detecting instruments is mainly addressed by approaches through classifying images according to their tool presence \cite{Twinanda2016}, which differs from usual object detection and localization tasks. This task, on the other hand, has recently been studied by using region-based Convolutional Neural Network (R-CNN) for learning instrument regions \cite{Jin2018} in cholecystectomy. The detection rate in both tasks, tool presence and spatial localization alike, has been improved by the usage of deep learning. 

Several approaches have been proposed for robotic instrument segmentation, ranging from watershed transformations~\cite{Lee2019SegmentationRI} over Fully Convolutional Networks (FCNs)~\cite{Long2015FCN} to U-Net\cite{Ronneberger2015UNet} architectures and beyond that combinations and modifications~\cite{Lee2019SegmentationRI, Pakhomov2017RIResNetFCN, Garcia2017ToolNet, Laina2017}. However, this task aims only to segment images into semantically interesting parts to answer the question of which image pixels belong to which class like instrument shaft or manipulator. It does not answer how many instances of this class are visible or rather if instances are occluded or not. Instance segmentation, on the other hand, requires an additional step to separate individual objects from each other and this is mainly addressed by subsequent pipeline stages concatenating semantic segmentation and tracking approaches \cite{Garcia2017SIFCN, Garcia2017ToolNet, Laina2017}.

Although semantic image segmentation has been studied extensively in the last years and various approaches have been proposed, no one to the best of our knowledge has studied the task of multi-class instance segmentation and recognition of instruments in gynecologic laparoscopy using a deep learning instance segmentation approach.

\section{Methodology}
We approach the task of instrument segmentation and recognition in laparoscopy by employing the Mask~R-CNN~\cite{He2017MaskR-CNN} that uses a region-based Convolutional Neural Network (CNN) as a basis, as well as a custom generated dataset, designed for segmenting multiple instruments in videos of laparoscopic gynecology. Furthermore, we consider different data augmentation techniques for this domain and evaluate the performance of the trained models in two different ways: (1) binary instrument segmentation to distinguish between instrument instance and background without recognizing the actual instrument, and (2) multi-class instrument recognition.

\subsection{Network Architecture}
\label{subsec:network-architecture}

Mask~R-CNN~\cite{He2017MaskR-CNN} can be described as region-based Fully Convolutional Network (R-FCN), which combines two types of network modules: a Region Proposal Network~(RPN)~\cite{Ren2017FasterR-CNN} and a Fully Convolutional Network~(FCN)~\cite{Long2015FCN}. It is an extension of Faster R-CNN~\cite{Ren2017FasterR-CNN}, a network targeted at multiple object detection in images, able to identify class affiliations of interesting regions by learning positions of objects. For a given input image, the output of Mask R-CNN is a set of region proposals comprising their bounding box coordinates in the input image, the probability of each object class as well as the segmentation mask for the object in this region. The idea of Mask R-CNN is that it uses the proposed regions from Faster R-CNN together with their class probabilities in order to separate these regions from its background. Therefore, the FCN module in the R-FCN is only used to separate pixels from its background without determining its class affiliations because this is provided by the previous module.

As a backbone network, we propose ResNet with 101 convolutional layers (ResNet-101), since this network has been successfully applied for the semantic segmentation task in robotic surgery \cite{Pakhomov2017RIResNetFCN}. The entire network is trainable end-to-end by adding up three loss functions: (1) loss of classification, (2) loss of localization and (3) average binary cross-entropy loss. Finally, the latter loss function is targeted at producing segmentation masks for an already classified region.

\subsection{Dataset}
\label{subsec:dataset}
In the field of medical endoscopy, only one dataset exist containing segmentation mask annotations for instruments from surgery videos. However, this dataset supports only the task of semantic segmentation in robot-assisted surgeries, which means they provide segmentation masks for different parts of specific robotic instruments. Since we focus on instrument segmentation in gynecology, and robotic tools differ from instruments in traditional laparoscopy, we introduce a custom dataset, acquired from various surgical interventions in gynecologic myomectomy and hysterectomy. 

\begin{figure*}[ht!]

    \subfloat[]{%
    \includegraphics[width=2.15cm]{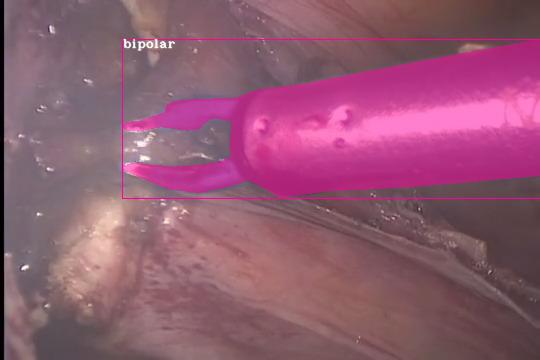}
    \label{fig:c}
    }
    \subfloat[]{%
    \includegraphics[width=2.15cm]{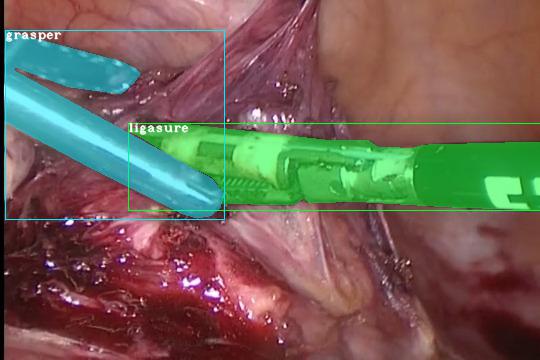}
    \label{fig:b}
    }
    \subfloat[]{%
    \includegraphics[width=2.15cm]{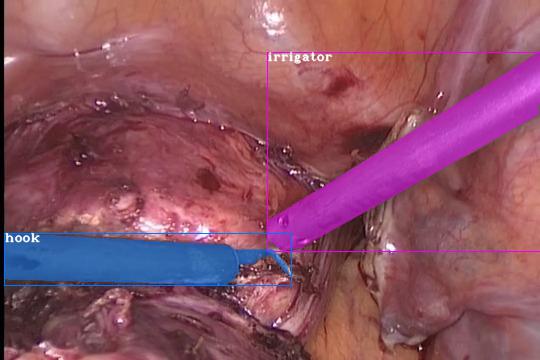}
    \label{fig:a}
    }
    \subfloat[]{%
    \includegraphics[width=2.15cm]{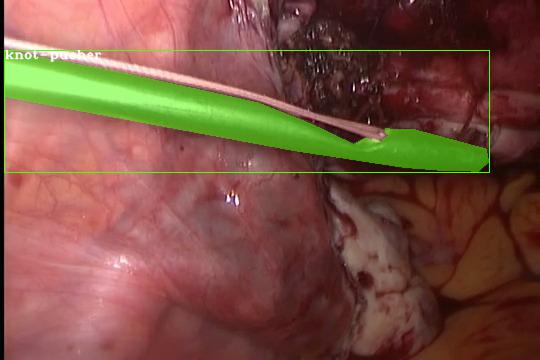}
    \label{fig:f}
    }
    \subfloat[]{%
    \includegraphics[width=2.15cm]{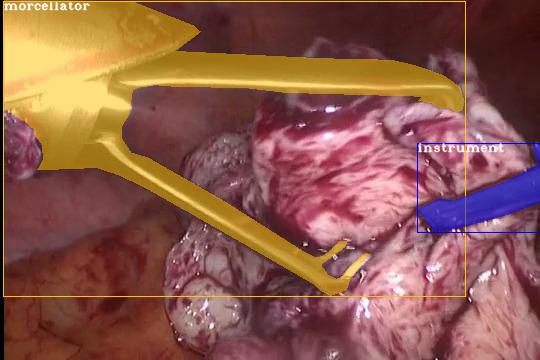}
    \label{fig:g}
    }
    \subfloat[]{%
    \includegraphics[width=2.15cm]{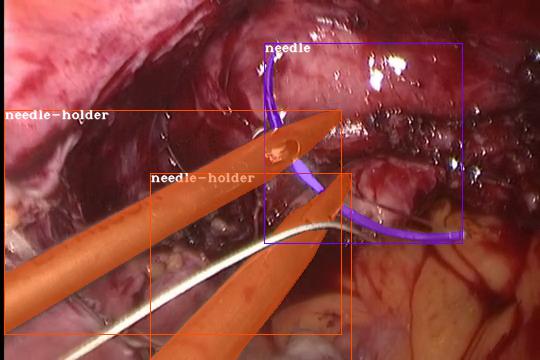}
    \label{fig:e}
    }
    \subfloat[]{%
    \includegraphics[width=2.15cm]{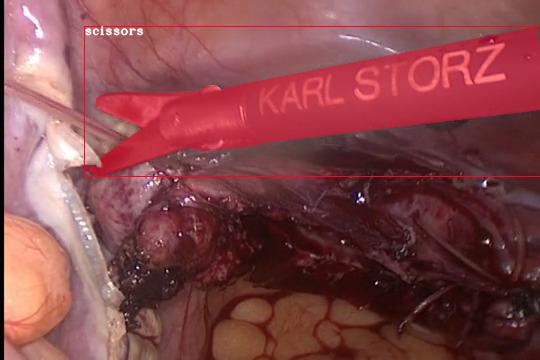}
    \label{fig:d}
    }
    \subfloat[]{%
    \includegraphics[width=2.15cm]{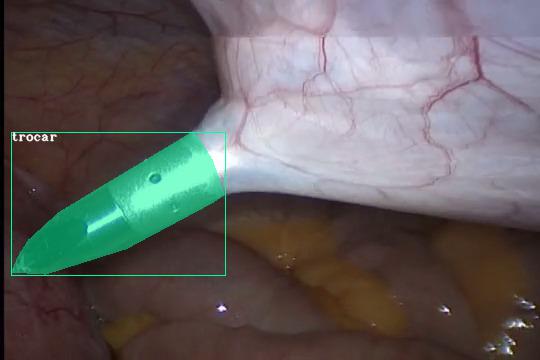}
    \label{fig:h}
  }
  \caption[]{Instruments and segmentation masks: (a) Bipolar, (b) Grasper (left), Sealer and Divider (right), (c) Hook (left), Irrigator (right), (d) Knot-Pusher, (e) Morcellator (left), Other (right), (f) Needle (center), Needle-Holders (left), (g) Scissors and (h) Trocar}
  \label{fig:dataset}
\end{figure*}

In total, the dataset consists of 333 randomly selected video frames with a resolution of $540\times360$, taken from videos of several different  laparoscopic surgeries (performed by a few different surgeons). These frames have been manually segmented according to their visible instruments, which results into 561 segmentation masks for different instruments. Ground-truth examples are shown in Figure~\ref{fig:dataset} and Table~\ref{tab:dataset} details the distribution of segmented instruments. As can be seen, we identified 11 different types of instruments, namely energy devices like \textit{Bipolar Grasper} and \textit{Hook} as well as \textit{Sealer} and \textit{Divider}. Furthermore, there are \textit{Grasper}, \textit{Irrigator}, \textit{Knot-Pusher}, \textit{Needle-Holder} and \textit{Scissors}. However, instruments like \textit{Morcellator}, \textit{Needle} and \textit{Trocar} are exceptions since these  instrument types differ strongly from others in appearance. The last type ``\textit{Other}'' in Table~\ref{tab:dataset} highlights instruments that are not identifiable due to occlusion caused by other instruments or tissue or due to distorted appearance. Since several instruments are visible in one image, the resulting number of images is smaller than the total obtained segmentation mas.

\begin{table}[hb!]
\centering
\begin{filecontents*}{tables/dataset.csv}
Instrument type;Instance samples;Augmented sample size
(a) Bipolar;38;311
(b) Grasper;60;259
(c) Hook;36;154
(c) Irrigator;39;282
(d) Knot-Pusher;37;526
(e) Morcellator;38;199
(f) Needle;55;440
(f) Needle-Holder;53;390
(g) Scissors;36;288
(b) Sealer and Divider;37;274
(h) Trocar;42;738
(e) Other;90;479
Total instances;561;4,340
Images;333;3,274
\end{filecontents*}
\pgfplotstabletypeset[
      col sep=semicolon,
      string type,
      assign column name/.style={/pgfplots/table/column name={\textbf{#1}}},
      columns/Class/.style={column name=Classification, column type={l}},
      every head row/.style={before row=\toprule, after row=\midrule},
      every last row/.style={after row=\hline},
    display columns/0/.style={column type={p{2.5cm}},reset styles,string type},
    every last row/.style={after row=\bottomrule},
    every row no 12/.style={before row={\toprule}, after row=\midrule}
    ]{tables/dataset.csv}
\caption{Details of the dataset: Showing the number of segmented instances per instrument type, distributed over 333 images as well as the resulting augmented sample size.}
\label{tab:dataset}
\end{table}

\subsection{Data Augmentation}
\label{subsec:data-augmenation}
To train a model based on the Mask R-CNN architecture, described in \ref{subsec:network-architecture}, our dataset includes only few training data, therefore the sample size of the dataset is additionally augmented by affine transformations and blurring. We consider only those transformations that preserve our segmentation labels such as (i) rotation, (ii) scaling, (iii) translation and (iv) mirroring. To detail the augmentation techniques, images are rotated at an angle of $\alpha \in \{0,45,90,135,180,225,270,315\}$, scaled with a factor of $c \in \{1.25,1.50,1.75\}$ and translated along the x- and y-axis $(u,v) \in \mathbb{R}^2$, where $u, v \in \{-0.1,0.1\}$, moving the image 10\% of their width and height along each axis. For all these transformations, new pixels are filled up by the average RGB value of the corresponding image, because studies have shown \cite{Obeso2019_DataAugm} that this approach works pretty reliable. Finally, we use Gaussian blur with $\sigma \in [0,3.0]$, as well as mirroring along the x-axis and y-axis.

Since not only one of these techniques can be applied to an image but also combinations of them like rotation and scaling together with blurring and mirroring, we propose two different augmentation stages: beforehand (offline) and on the fly during training (online). Firstly, images are processed offline, where different rotations, scales, and translations are applied. We introduce a threshold that skips transformations that leads to loss of labels and keep only augmented images where all instrument labels remain. We compare the area of the original segmentation mask to the augmented mask and keep only transformations that differ slightly from its original, measured by the size of area. During a training phase, data are additionally augmented by using blur and flipping randomly with a probability of 50\% to the input image sequence. 

\subsection{Evaluation Criteria}
\label{subsec:evaluation-criteria}

For evaluating the performance of trained models quantitatively, we use proposed COCO metrics~\cite{Lin2015_COCO} as evaluation criteria. These metrics are based on average precision and recall and are employed in the COCO competition to measure the performance of object detection and instance segmentation algorithms. A positive detection or segmentation is counted as true positive whenever a predicted and true instance is similar according to their bounding boxes or segmentation masks. This similarity is measured by the Intersection over Union (IoU), also referred to as Jaccard index and is defined as $IoU = \frac{|T \cap D|}{|T \cup D|}$. It describes the ratio of the overlapping region between the area of ground-truth $T$ and the detected area $D$ to their total area, which is calculated for each instance per image. Performance measures using COCO metrics comprises average precision values for the entire dataset over different IoU thresholds and are defined as AP$_{50:95}$ and AP$_{50}$: Whereas AP$_{50}$ is the average precision with IoU threshold of 50\% and AP$_{50:95}$ is additionally averaged over IoU thresholds from 50\% to 90\% obtained in steps of 5\%. Furthermore, average recall is specified as AR and describes the average recall of detections according to the number of maximal instances per image.

\section{Experiments and Results}
In this section, we describe in detail the experimental setup used to train and evaluate models based on Mask~R-CNN for the task of instrument segmentation and recognition in laparoscopy. We focus on two scenarios: binary instance segmentation as well as multi-class instance segmentation and recognition. Whereas binary segmentation targets at identifying instrument instances only, which means that instrument instances are separated from its background, multi-class segmentation and recognition mean that instruments are not only segmented but also classified according to its type.

\subsection{Implementation Details}

As described in Section \ref{subsec:dataset}, we use a custom dataset designed for the task of multi-class instrument recognition in laparoscopy videos. To train models with different setups, we split the entire dataset into three subsets. We use 60\% of the images for training and 20\% for the validation of the training phase in order to prevent overfitting. The remaining 20\% of the images are used to evaluate the prediction performance of trained models after each epoch. Furthermore, we consider that each split consists of an equal number of instances per instrument type, which results in seven examples per instrument class in the validation and test split, respectively. Moreover, we apply data augmentation techniques only to the training dataset and keep validation and test set unchanged in order to compare different setups.

Our experiments are conducted using a publicly available implementation of Mask R-CNN\footnote{\url{https://github.com/matterport/Mask_RCNN}}. Since our dataset is small even with data augmentation techniques, we have compared training from scratch with transfer learning as a first setup. Especially the latter is compared by fine-tuning the last layers and further training of all layers. For initializing the model in this setup, we use pre-trained weights on the COCO dataset\cite{Lin2015_COCO}. Although this data strongly differs from images in the medical domain, using pre-trained weights obtained by this data as initialization leads to faster and more stable training phases. However, it has shown that further training all layers of the network is more beneficial for our data than only fine-tuning the last layers of the classifier. To setup the network, we use proposed hyper-parameters in \cite{He2017MaskR-CNN} and apply Stochastic Gradient Descent (SDG) as optimizer. Since it is not clear which learning rate is beneficial for our dataset, we have compared different learning rates in the range of $\eta \in \{0.01, 0.001, 0.0001\}$ with momentum $\mu$=0.9 and the optimal learning rate for all loss function is measured with a constant learning rate of $\eta$= 0.001. 

\subsection{Quantitative \& Qualitative Results}
\label{subsec:quantitative-results}

As a baseline setup for comparison, we leverage the proposed network from Section~\ref{subsec:network-architecture} to train our custom dataset (see \ref{subsec:dataset}) for a binary instrument segmentation task. In a first setup, we investigate the prediction performance of the network to detect and segment instruments without identifying its instrument type using the data augmentation techniques described in Section~\ref{subsec:data-augmenation}. Table~\ref{tab:ap_bbox_mask} summarizes the performance after each epoch and as can be seen, average precision of segmentation masks comprising different IoU thresholds (AP$_{50:95}$) increases faster with the augmented sample size during training but also time for training is increased by a factor of twelve per epoch. However, taking a closer look at the AP with IoU threshold of 50\% (AP$_{50}$), it becomes clear that both approaches perform similarly good and achieve an average precision of approx. 82\%, thus, data augmentation causes only an increase in training time for the binary instrument segmentation task. The localization of instruments (see Table~\ref{tab:ap_bbox_mask} bounding boxes (baseline)), on the other hand, achieves its highest average precision already after the first five epochs for the IoU thresholds of 50\% (AP$_{50}$). However, the average precision of bounding box predictions comprising different IoU thresholds (AP$_{50:95}$) is similar to the average precision of predicted segmentation masks, namely at epoch 30 with an average precision of 64,50\% compared to 60,06\% without an augmented sample size. 

\begin{table*}[ht!]
	\centering
	\begin{filecontents*}{tables/ap_bbox_mask_all.csv}
Setup,Measure,5,10,15,20,25,30,35,40,45,50
\textbf{Segmentation masks (baseline)},,,,,,,,,,,
Binary instrument,AP$_{50:95}$,0.414,0.432,0.429,0.523,0.503,0.522,0.499,\textbf{0.531},0.528,0.527
Binary instrument,AP$_{50}$,0.700,0.696,0.698,0.794,0.774,\textbf{0.820},0.751,0.806,0.808,0.805
Binary instrument (augmented),AP$_{50:95}$,0.500,0.454,0.478,0.518,0.509,\textbf{0.543},0.520,0.500,0.473,0.510
Binary instrument (augmented),AP$_{50}$,0.807,0.737,0.761,0.811,0.767,\textbf{0.814},0.795,0.750,0.726,0.795
Multi-class instrument,AP$_{50:95}$,0.046,0.067,0.163,0.150,0.203,0.258,0.188,0.250,0.330,\textbf{0.331}
Multi-class instrument,AP$_{50}$,0.085,0.157,0.320,0.287,0.376,0.411,0.333,0.409,\textbf{0.526},0.511
Multi-class instrument (augmented),AP$_{50:95}$,0.178,0.146,0.226,0.350,0.365,0.398,0.413,0.375,0.389,\textbf{0.429}
Multi-class instrument (augmented),AP$_{50}$,0.296,0.270,0.384,0.523,0.579,0.590,0.604,0.525,0.557,\textbf{0.613}
\textbf{Bounding boxes (baseline)},,,,,,,,,,,
Binary instrument,AP$^{bb}_{50:95}$,0.471,0.518,0.522,0.579,0.613,0.606,0.558,0.614,0.627,\textbf{0.631}
Binary instrument,AP$^{bb}_{50}$,0.777,0.806,0.741,0.820,0.827,\textbf{0.833},0.805,0.804,0.820,0.816
Binary instrument (augmented),AP$^{bb}_{50:95}$,0.628,0.572,0.593,0.642,0.633,\textbf{0.645},0.629,0.584,0.557,0.634
Binary instrument (augmented),AP$^{bb}_{50}$,\textbf{0.865},0.793,0.810,0.851,0.826,0.831,0.826,0.763,0.747,0.832
Multi-class instrument,AP$^{bb}_{50:95}$,0.046,0.072,0.169,0.17,0.208,0.261,0.197,0.266,\textbf{0.354},0.339
Multi-class instrument,AP$^{bb}_{50}$,0.087,0.186,0.343,0.343,0.367,0.450,0.332,0.413,\textbf{0.561},0.532
Multi-class instrument (augmented),AP$^{bb}_{50:95}$,0.188,0.168,0.241,0.316,0.364,0.419,0.444,0.397,0.412,\textbf{0.457}
Multi-class instrument (augmented),AP$^{bb}_{50}$,0.326,0.259,0.378,0.527,0.561,0.598,0.600,0.554,0.554,\textbf{0.627}
\end{filecontents*}
\pgfplotstabletypeset[
      col sep=comma,
      string type,
      assign column name/.style={/pgfplots/table/column name={\textbf{#1}}},
      every head row/.style={before row=\toprule},
      every last row/.style={after row=\hline},
      display columns/0/.style={column type={p{4.2cm}|},reset styles,string type},
      display columns/1/.style={column type={p{1.0cm}|}},
      display columns/2/.style={column type={p{0.8cm}|}},
      display columns/3/.style={column type={p{0.8cm}|}},
      display columns/4/.style={column type={p{0.8cm}|}},
      display columns/5/.style={column type={p{0.8cm}|}},
      display columns/6/.style={column type={p{0.8cm}|}},
      display columns/7/.style={column type={p{0.8cm}|}},
      display columns/8/.style={column type={p{0.8cm}|}},
      display columns/9/.style={column type={p{0.8cm}|}},
      display columns/10/.style={column type={p{0.8cm}|}},
      every row no 0/.style={before row={\midrule}, after row=\midrule},
      every row no 3/.style={before row={\hline}},
      every row no 5/.style={before row={\hline}},
      every row no 7/.style={before row={\hline}},
      every row no 9/.style={before row={\midrule}, after row=\midrule},
      every row no 12/.style={before row={\hline}},
      every row no 14/.style={before row={\hline}},
      every row no 16/.style={before row={\hline}},
      every last row/.style={after row=\bottomrule},
    ]{tables/ap_bbox_mask_all.csv}
	\caption{Average precision of predicted \textbf{segmentation masks AP} and \textbf{bounding boxes AP$^{bb}$} for binary instrument segmentation and multi-class instrument recognition, evaluated on test split with different IoU thresholds after each epoch.}
	\label{tab:ap_bbox_mask}
\end{table*}

\begin{figure}[ht!]
    \subfloat[Overall loss]{
    \includegraphics[width=\linewidth,trim=0.2cm 0.5cm 0.0cm 2.0 cm,clip]{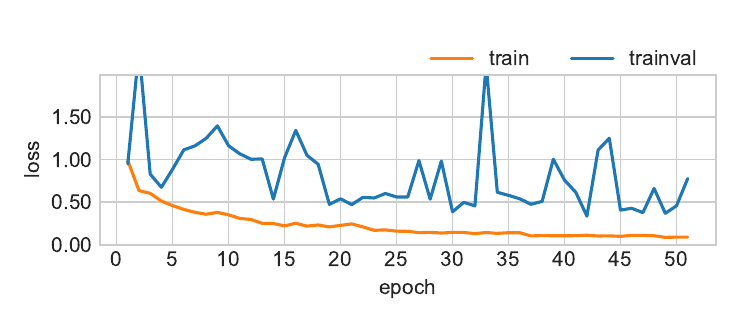}
    \label{fig:loss_b_a}
  }\\
    \subfloat[Average precision with IoU threshold of 50\% (AP$_{50}$)]{%
    \includegraphics[width=\linewidth,trim=0.0cm 0.0cm 0.0cm 0.0 cm,clip]{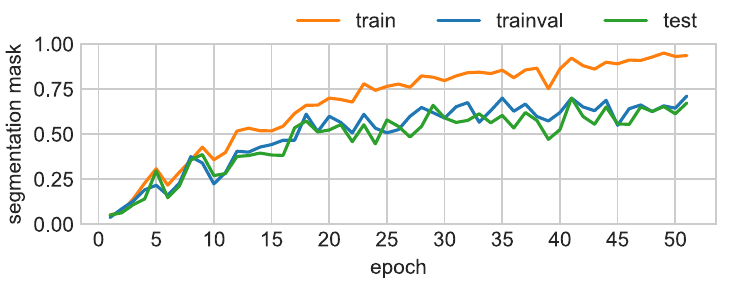}
    \label{fig:ap_b_a}
  }
  \caption[]{Loss and average precision during training the network for the \textbf{multi-class segmentation and recognition task} using the augmented sample size.}
  \label{fig:training_multi-class_augm}
\end{figure}

Since we focus on multi-class instrument segmentation and recognition, we employ the same setup to train a model for identifying eleven different classes of instruments plus one default class that comprises unspecified instruments. In comparison to the binary instrument segmentation, this task is more challenging because some instruments look similar to others and only few samples are included in the dataset to distinguish between them. The results for the multi-class classification are summarized in Table~\ref{tab:ap_bbox_mask} and although these results are much lower in comparison to the binary segmentation, data augmentation improves the average precision for this task at both IoU thresholds (AP$_{50:95}$, AP$_{50}$) and achieves an average precision of approx. 62\% with IoU threshold of 50\% (AP$_{50}$). Figure~\ref{fig:training_multi-class_augm} details the performance during training and shows overall loss and average precision with IoU threshold of 50\% (AP$_{50}$). As can be seen, the validation loss (see Figure~\ref{fig:training_multi-class_augm}(a) blue line) during training fluctuates and is slightly higher than the training loss but validation loss still decreases over time. However, after the 50th epoch validation loss increases rapidly and further training leads to overfitting.

Finally, Table~\ref{tab:multi-class-details} details the classification performance for each instrument at the 50th epoch using the augmented sample size. It details the average precision as well as the average recall of the trained model for predicting segmentation masks and bounding boxes, respectively. A high recall (AR$_{1}$) indicates that instruments like Hook, Knot-Pusher, Needle and Trocar are more often classified than other instruments. When taking a closer look at average precision between segmentation masks and bounding boxes, remarkable differences can be regarded between instruments like Bipolar Grasper and Knot-Pusher, reflecting the fact that these instruments have more complex segmentation masks. Furthermore, there are specific instruments which are localized with higher precision such as Bipolar, Hook, Irrigator as well as Sealer and Divider. However, the prediction performance of segmentation masks decreases minimally for all instruments in comparison to their localization. With a threshold of 50\%, less than the half of instruments are recognized with reasonable precision and instruments such as Grasper, Knot-Pusher, Needle-Holder, Scissors seems to be difficult to distinguish.

\begin{figure*}[ht!]
    \centering
    \includegraphics[width=14.90cm,trim=0.0cm 0.0cm 0.0cm 2.0 cm,clip]{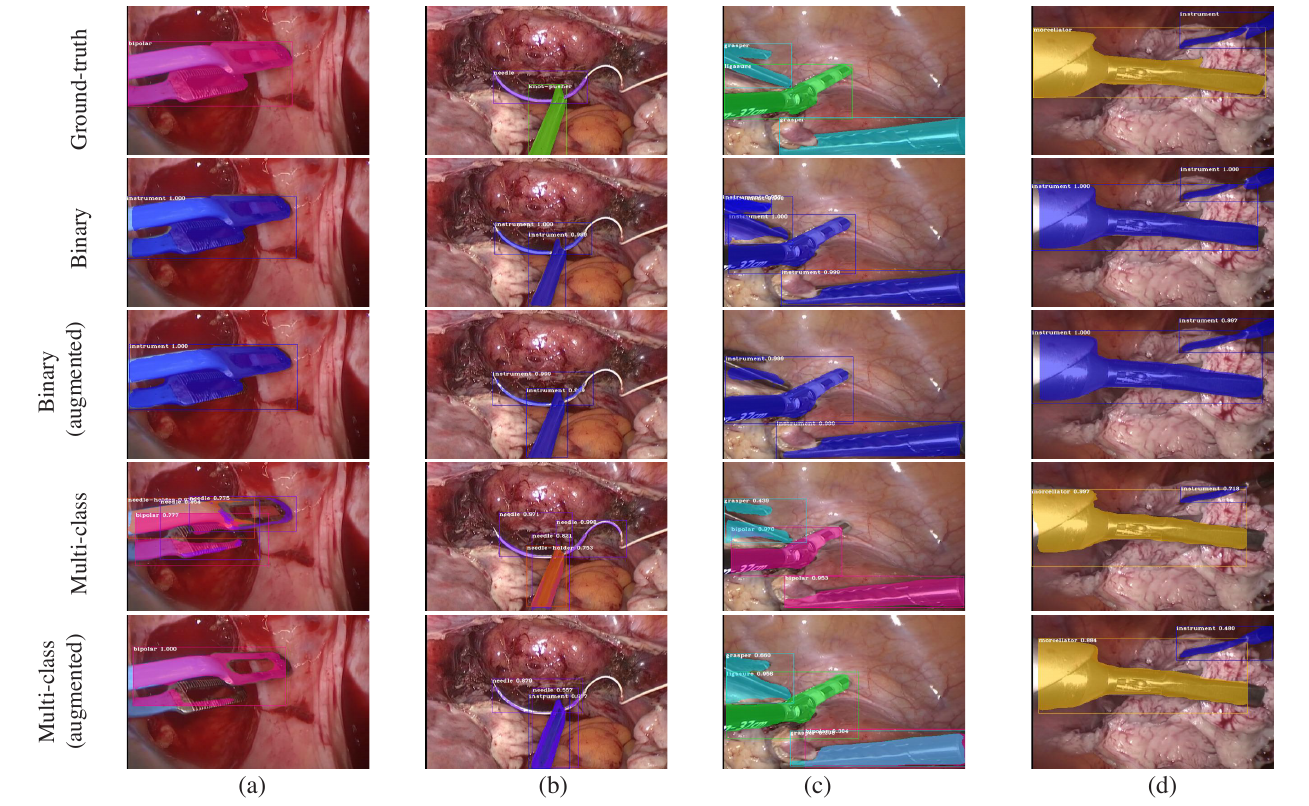}
    \caption{Qualitative results showing segmentation of four test images: (a) Needle-Holder (center), Needle (center) (b) Bipolar, (c) Grasper (right \& left), Sealer and Divider (left), (d) Morcellator (left), Other (right). The top row illustrates ground-truth segmentation and remaining one shows automatic predicted binary and multi-class segmentation masks compared with and without data augmentation.}
    \label{fig:qualitative-results}
\end{figure*}

Some qualitative comparisons for both segmentation tasks comparing additionally data augmentation are illustrated in Figure~\ref{fig:qualitative-results}. The top row shows the ground-truth segmentation mask compared to the binary segmentation and multi-class recognition as well as with and without data augmentation. As already stated previously, augmentation for the binary segmentation task does not improve results for the segmentation mask, however, it is beneficial for the multi-class recognition task. Furthermore, Figure~\ref{fig:qualitative-results}(a) demonstrates the difficult to segment more complex masks for instruments like Bipolar. Also, needles (see Figure~\ref{fig:qualitative-results}(b)) are difficult to segment because they are thin tools and hard to distinguish from its background but they are partially detected. However, the figure underlines the difficulty in separating instruments considering multi-class classifications, whereas results for the binary segmentation task are reasonably accurate to use it as input for further analysis methods.

\begin{table}[ht!]
\centering
\begin{filecontents*}{tables/detail_best_settings.csv}
Instruments,AP$_{50:95}$,AP$_{50}$,AR$_{1}$,AP$^{bb}_{50:95}$,AP$^{bb}_{50}$,AR$^{bb}_{1}$
Bipolar,0.392,\textbf{0.851},0.471,\textbf{0.655},\textbf{1.000},\textbf{0.700}
Grasper,0.417,0.55,0.3,0.469,0.55,0.386
Hook,\textbf{0.643},\textbf{0.812},\textbf{0.686},\textbf{0.635},\textbf{0.812},\textbf{0.686}
Irrigator,\textbf{0.756},\textbf{0.869},0.343,\textbf{0.710},\textbf{0.869},0.329
Knot-Pusher,0.387,0.574,\textbf{0.871},0.474,0.574,\textbf{0.829}
Morcellator,0.504,\textbf{0.791},0.4,0.507,\textbf{0.791},0.486
Needle,0.051,0.218,\textbf{0.643},0.055,0.236,\textbf{0.657}
Needle-Holder,0.333,0.517,0.057,0.331,0.517,0.1
Scissors,0.421,0.574,0.3,0.474,0.574,0.329
Sealer-Divider,\textbf{0.831},\textbf{1.000},0.429,\textbf{0.776},\textbf{1.000},0.486
Trocar,0.322,0.49,\textbf{0.843},0.306,0.49,\textbf{0.800}
Other,0.092,0.113,0.386,0.089,0.113,0.386
\textbf{Mean},\textbf{0.429},\textbf{0.613},\textbf{0.477},\textbf{0.457},\textbf{0.627},\textbf{0.514}
\textbf{Instrument},\textbf{0.543},\textbf{0.814},\textbf{0.327},\textbf{0.645},\textbf{0.831},\textbf{0.394}
\end{filecontents*}
\pgfplotstabletypeset[
      col sep=comma,
      string type,
      assign column name/.style={/pgfplots/table/column name={\textbf{#1}}},
      columns/Class/.style={column name=Classification, column type={l}},
      every head row/.style={before row={\midrule & \multicolumn{3}{c|}{\textbf{Segmentation mask}} & \multicolumn{3}{c}{\textbf{Bounding box}}\\}},
      display columns/1/.style={column type={p{0.6cm}}},
      display columns/3/.style={column type={p{0.6cm}|}},
      display columns/4/.style={column type={p{0.6cm}}},
      display columns/6/.style={column type={p{0.6cm}}},
      every row no 12/.style={before row={\toprule}, after row=\bottomrule},
      every first row/.style={before row=\midrule},
      every last row/.style={after row=\hline},
    display columns/0/.style={column type={p{1.7cm}},reset styles,string type},
    every last row/.style={before row=\midrule, after row=\bottomrule},
    ]{tables/detail_best_settings.csv}
\caption{Quantitative results of the \textbf{multi-class} segmentation and recognition, detailed for each class using the best average precision (AP$_{50}$) obtained by the augmented sample size at the 50th epoch.}
\label{tab:multi-class-details}
\end{table}

\section{Conclusions}
In this work, we have evaluated a region-based fully convolutional network for instrument segmentation and recognition in videos recorded from  laparoscopic gynecology. Even though we use only a moderately small number of training examples, we have shown that the segmentation task works reliable, achieving an average precision of $\geq 81\%$ (at a minimum region overlap of 50\%) for both the binary and the multi-class classification approach. Moreover, we have shown that we are able to classify particular instruments in the proposed regions very accurately too (with 100\% AP for some instruments), but the performance heavily varies among different instruments.    

\section*{Acknowledgments}
This work was funded by the FWF Austrian Science Fund under grant P 32010-N38.


\bibliographystyle{ieeetran}



\end{document}